\documentclass{article}

\usepackage[final,nonatbib]{nips_2016}

\usepackage[utf8]{inputenc} 
\usepackage[T1]{fontenc}    
\usepackage{hyperref}       
\usepackage{url}            
\usepackage{booktabs}       
\usepackage{amsfonts}       
\usepackage{nicefrac}       
\usepackage{microtype}      
\usepackage{amsmath}
\usepackage{graphicx}
\usepackage{subcaption}
\usepackage{url}

\usepackage{xcolor}
\usepackage{listings}

\title{An Overview of Multi-Task Learning\\in Deep Neural Networks\thanks{This paper originally appeared as a blog post at \url{http://sebastianruder.com/multi-task/index.html} on 29 May 2017.}}

\author{
  Sebastian Ruder\\
  Insight Centre for Data Analytics, NUI Galway\\
  Aylien Ltd., Dublin\\
  \texttt{ruder.sebastian@gmail.com}
}

\begin{document}

\maketitle

\begin{abstract}

Multi-task learning (MTL) has led to successes in many applications of machine learning, from natural language processing and speech recognition to computer vision and drug discovery. This article aims to give a general overview of MTL, particularly in deep neural networks. It introduces the two most common methods for MTL in Deep Learning, gives an overview of the literature, and discusses recent advances. In particular, it seeks to help ML practitioners apply MTL by shedding light on how MTL works and providing guidelines for choosing appropriate auxiliary tasks.

\end{abstract}

\section{Introduction}

In Machine Learning (ML), we typically care about optimizing for a particular metric, whether this is a score on a certain benchmark or a business KPI. In order to do this, we generally train a single model or an ensemble of models to perform our desired task. We then fine-tune and tweak these models until their performance no longer increases. While we can generally achieve acceptable performance this way, by being laser-focused on our single task, we ignore information that might help us do even better on the metric we care about. Specifically, this information comes from the training signals of related tasks. By sharing representations between related tasks, we can enable our model to generalize better on our original task. This approach is called Multi-Task Learning (MTL).

Multi-task learning has been used successfully across all applications of machine learning, from natural language processing \cite{Collobert2008} and speech recognition \cite{Deng2013a} to computer vision \cite{Girshick2015} and drug discovery \cite{Ramsundar2015}. MTL comes in many guises: joint learning, learning to learn, and learning with auxiliary tasks are only some names that have been used to refer to it. Generally, as soon as you find yourself optimizing more than one loss function, you are effectively doing multi-task learning (in contrast to single-task learning). In those scenarios, it helps to think about what you are trying to do explicitly in terms of MTL and to draw insights from it.

Even if you are only optimizing one loss as is the typical case, chances are there is an auxiliary task that will help you improve upon your main task. \cite{Caruana1998} summarizes the goal of MTL succinctly: ``MTL improves generalization by leveraging the domain-specific information contained in the training signals of related tasks".

Over the course of this article, I will try to give a general overview of the current state of multi-task learning, in particular when it comes to MTL with deep neural networks. I will first motivate MTL from different perspectives in Section \ref{sec:motivation}. I will then introduce the two most frequently employed methods for MTL in Deep Learning in Section \ref{sec:two-methods}. Subsequently, in Section \ref{sec:why_does_it_work}, I will describe mechanisms that together illustrate why MTL works in practice. Before looking at more advanced neural network-based MTL methods, I will provide some context in Section \ref{sec:literature} by discussing the literature in MTL. I will then introduce some more powerful recently proposed methods for MTL in deep neural networks in Section \ref{sec:recent_work}. Finally, I will talk about commonly used types of auxiliary tasks and discuss what makes a good auxiliary task for MTL in Section \ref{sec:auxiliary_tasks}.

\section{Motivation} \label{sec:motivation}

We can motivate multi-task learning in different ways: Biologically, we can see multi-task learning as being inspired by human learning. For learning new tasks, we often apply the knowledge we have acquired by learning related tasks. For instance, a baby first learns to recognize faces and can then apply this knowledge to recognize other objects.

From a pedagogical perspective, we often learn tasks first that provide us with the necessary skills to master more complex techniques. This is true for learning the proper way of falling in martial arts, e.g. Judo as much as learning to program.

Taking an example out of pop culture, we can also consider \emph{The Karate Kid} (1984)\footnote{Thanks to Margaret Mitchell and Adrian Benton for the inspiration}. In the movie, \emph{sensei} Mr Miyagi teaches the karate kid seemingly unrelated tasks such as sanding the floor and waxing a car. In hindsight, these, however, turn out to equip him with invaluable skills that are relevant for learning karate.

Finally, we can motivate multi-task learning from a machine learning point of view: We can view multi-task learning as a form of inductive transfer. Inductive transfer can help improve a model by introducing an inductive bias, which causes a model to prefer some hypotheses over others. For instance, a common form of inductive bias is $\ell_1$ regularization, which leads to a preference for sparse solutions. In the case of MTL, the inductive bias is provided by the auxiliary tasks, which cause the model to prefer hypotheses that explain more than one task. As we will see shortly, this generally leads to solutions that generalize better.

\section{Two MTL methods for Deep Learning} \label{sec:two-methods}

So far, we have focused on theoretical motivations for MTL. To make the ideas of MTL more concrete, we will now look at the two most commonly used ways to perform multi-task learning in deep neural networks. In the context of Deep Learning, multi-task learning is typically done with either \emph{hard} or \emph{soft parameter sharing} of hidden layers.

\begin{figure}[!htb]
      \centering
         \includegraphics[width=0.5 \linewidth]{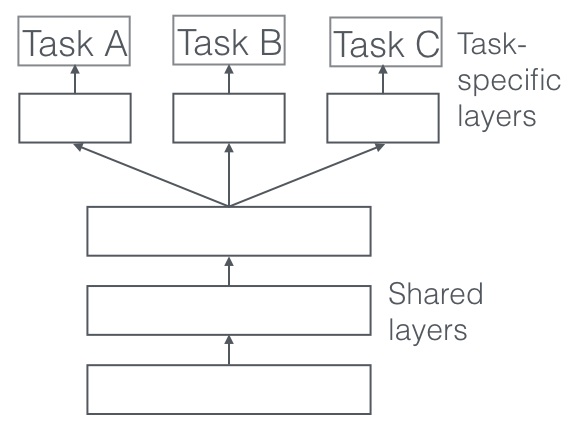}
	\caption{Hard parameter sharing for multi-task learning in deep neural networks}
	\label{fig:hard_parameter_sharing}
\end{figure}

\subsection{Hard parameter sharing}

Hard parameter sharing is the most commonly used approach to MTL in neural networks and goes back to \cite{Caruana1993}. It is generally applied by sharing the hidden layers between all tasks, while keeping several task-specific output layers as can be seen in Figure \ref{fig:hard_parameter_sharing}.

Hard parameter sharing greatly reduces the risk of overfitting. In fact, \cite{Baxter1997} showed that the risk of overfitting the shared parameters is an order $N$ -- where $N$ is the number of tasks -- smaller than overfitting the task-specific parameters, i.e. the output layers. This makes sense intuitively: The more tasks we are learning simultaneously, the more our model has to find a representation that captures all of the tasks and the less is our chance of overfitting on our original task.

\subsection{Soft parameter sharing}

\begin{figure}[!htb]
      \centering
         \includegraphics[width=0.8 \linewidth]{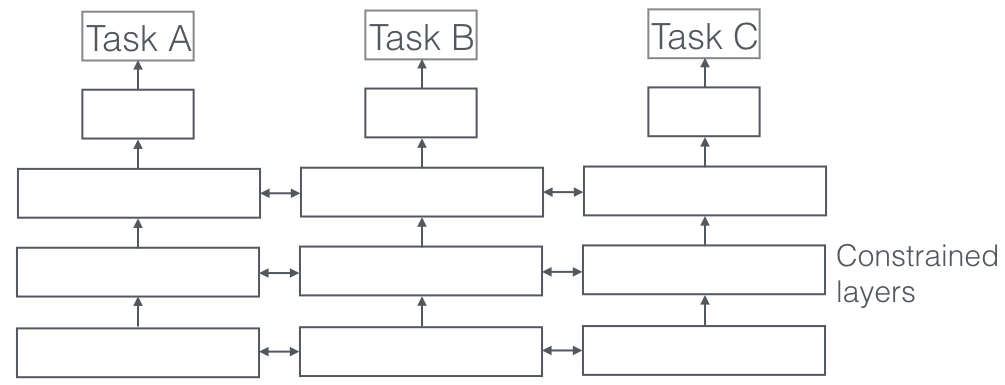}
	\caption{Soft parameter sharing for multi-task learning in deep neural networks}
	\label{fig:soft_parameter_sharing}
\end{figure}

In soft parameter sharing on the other hand, each task has its own model with its own parameters. The distance between the parameters of the model is then regularized in order to encourage the parameters to be similar, as evidenced in Figure \ref{fig:soft_parameter_sharing}. \cite{Duong2015a} for instance use $\ell_2$ distance for regularization, while \cite{Yang2017a} use the trace norm.

The constraints used for soft parameter sharing in deep neural networks have been greatly inspired by regularization techniques for MTL that have been developed for other models, which we will soon discuss.

\section{Why does MTL work?} \label{sec:why_does_it_work}

Even though an inductive bias obtained through multi-task learning seems intuitively plausible, in order to understand MTL better, we need to look at the mechanisms that underlie it. Most of these have first been proposed by \cite{Caruana1998}. For all examples, we will assume that we have two related tasks $A$ and $B$, which rely on a common hidden layer representation $F$. 

\subsection{Implicit data augmentation}

MTL effectively increases the sample size that we are using for training our model. As all tasks are at least somewhat noisy, when training a model on some task $A$, our aim is to learn a good representation for task $A$ that ideally ignores the data-dependent noise and generalizes well. As different tasks have different noise patterns, a model that learns two tasks simultaneously is able to learn a more general representation. Learning just task $A$ bears the risk of overfitting to task $A$, while learning $A$ and $B$ jointly enables the model to obtain a better representation $F$ through averaging the noise patterns.

\subsection{Attention focusing}

If a task is very noisy or data is limited and high-dimensional, it can be difficult for a model to differentiate between relevant and irrelevant features. MTL can help the model focus its attention on those features that actually matter as other tasks will provide additional evidence for the relevance or irrelevance of those features.

\subsection{Eavesdropping}

Some features $G$ are easy to learn for some task $B$, while being difficult to learn for another task $A$. This might either be because $A$ interacts with the features in a more complex way or because other features are impeding the model's ability to learn $G$. Through MTL, we can allow the model to \emph{eavesdrop}, i.e. learn $G$ through task $B$. The easiest way to do this is through \emph{hints} \cite{Abu-Mostafa1990}, i.e. directly training the model to predict the most important features.

\subsection{Representation bias}

MTL biases the model to prefer representations that other tasks also prefer. This will also help the model to generalize to new tasks in the future as a hypothesis space that performs well for a sufficiently large number of training tasks will also perform well for learning novel tasks as long as they are from the same environment \cite{Baxter2000}.

\subsection{Regularization}

Finally, MTL acts as a regularizer by introducing an inductive bias. As such, it reduces the risk of overfitting as well as the Rademacher complexity of the model, i.e. its ability to fit random noise.

\section{MTL in non-neural models} \label{sec:literature}

In order to better understand MTL in deep neural networks, we will now look to the existing literature on MTL for linear models, kernel methods, and Bayesian algorithms. In particular, we will discuss two main ideas that have been pervasive throughout the history of multi-task learning: enforcing sparsity across tasks through norm regularization; and modelling the relationships between tasks.

Note that many approaches to MTL in the literature deal with a homogenous setting: They assume that all tasks are associated with a single output, e.g. the multi-class MNIST dataset is typically cast as 10 binary classification tasks. More recent approaches deal with a more realistic, heterogeneous setting where each task corresponds to a unique set of outputs. 

\subsection{Block-sparse regularization}

\paragraph{Notation} In order to better connect the following approaches, let us first introduce some notation. We have $T$ tasks. For each task $t$, we have a model $m_t$ with parameters $a_t$ of dimensionality $d$. We can write the parameters as a column vector $a_t$:
$$a_t = \begin{bmatrix}a_{1, t} \\ \vdots \\ a_{d, t} \end{bmatrix}^\top$$
We now stack these column vectors $a_1, \ldots, a_T$ column by column to form a matrix $A \in 
\mathbb{R}^{d \times T}$. The $i$-th row of $A$ then contains the parameter $a_{i, \cdot}$ corresponding to the $i$-th feature of the model for every task, while the $j$-th column of $A$ contains the parameters $a_{\cdot,j}$ corresponding to the $j$-th model.

Many existing methods make some sparsity assumption with regard to the parameters of our models. \cite{Argyriou2007} assume that all models share a small set of features. In terms of our task parameter matrix $A$, this means that all but a few rows are $0$, which corresponds to only a few features being used across \emph{all} tasks. In order to enforce this, they generalize the $\ell_1$ norm to the MTL setting. Recall that the $\ell_1$ norm is a constraint on the sum of the parameters, which forces all but a few parameters to be exactly $0$. It is also known as lasso (\textbf{l}east \textbf{a}bsolute \textbf{s}hrinkage and \textbf{s}election \textbf{o}perator).

While in the single-task setting, the $\ell_1$ norm is computed based on the parameter vector $a_t$ of the respective task $t$, for MTL we compute it over our task parameter matrix $A$. In order to do this, we first compute an $\ell_q$ norm across each row $a_i$ containing the parameter corresponding to the $i$-th feature across all tasks, which yields a vector $b = \begin{bmatrix}\|a_1\|_q \ldots \|a_d\|_q \end{bmatrix} \in \mathbb{R}^d$. We then compute the $\ell_1$ norm of this vector, which forces all but a few entries of $b$, i.e. rows in $A$ to be $0$.

As we can see, depending on what constraint we would like to place on each row, we can use a different $\ell_q$. In general, we refer to these mixed-norm constraints as $\ell_1/\ell_q$ norms. They are also known as block-sparse regularization, as they lead to entire rows of $A$ being set to $0$. \cite{Zhang2008} use $\ell_1/\ell_\infty$ regularization, while \cite{Argyriou2007} use a mixed $\ell_1/\ell_2$ norm. The latter is also known as group lasso and was first proposed by \cite{Yuan2006a}.

\cite{Argyriou2007}  also show that the problem of optimizing the non-convex group lasso can be made convex by penalizing the trace norm of $A$, which forces $A$ to be low-rank and thereby constrains the column parameter vectors $a_{\cdot, 1}, \ldots, a_{\cdot, t}$ to live in a low-dimensional subspace. \cite{Lounici2009} furthermore establish upper bounds for using the group lasso in multi-task learning.

As much as this block-sparse regularization is intuitively plausible, it is very dependent on the extent to which the features are shared across tasks. \cite{Negahban2008} show that if features do not overlap by much, $\ell_1/\ell_q$ regularization might actually be worse than element-wise $\ell_1$ regularization.

For this reason, \cite{Jalali2010} improve upon block-sparse models by proposing a method that combines block-sparse and element-wise sparse regularization. They decompose the task parameter matrix $A$ into two matrices $B$ and $S$ where $A = B + S$. $B$ is then enforced to be block-sparse using $\ell_1/\ell_\infty$ regularization, while $S$ is made element-wise sparse using lasso. Recently, \cite{Liu2016g} propose a distributed version of group-sparse regularization.

\subsection{Learning task relationships}

While the group-sparsity constraint forces our model to only consider a few features, these features are largely used across all tasks. All of the previous approaches thus assume that the tasks used in multi-task learning are closely related. However, each task might not be closely related to all of the available tasks. In those cases, sharing information with an unrelated task might actually hurt performance, a phenomenon known as negative transfer. 

Rather than sparsity, we would thus like to leverage prior knowledge indicating that some tasks are related while others are not. In this scenario, a constraint that enforces a clustering of tasks might be more appropriate. \cite{Evgeniou2005} suggest to impose a clustering constraint by penalizing both the norms of our task column vectors $a_{\cdot, 1}, \ldots, a_{\cdot, t}$ as well as their variance with the following constraint:
$$\Omega = \|\bar{a}\|^2 + \dfrac{\lambda}{T} \sum^T_{t=1} \| a_{\cdot, t} - \bar{a} \|^2 $$
where $\bar{a} = (\sum^T_{t=1} a_{\cdot, t})/T $ is the mean parameter vector. This penalty enforces a clustering of the task parameter vectors $a_{\cdot, 1}, \ldots, a_{\cdot, t}$ towards their mean that is controlled by $\lambda$. They apply this constraint to kernel methods, but it is equally applicable to linear models.

A similar constraint for SVMs was also proposed by \cite{Evgeniou2004}. Their constraint is inspired by Bayesian methods and seeks to make all models close to some mean model. In SVMs, the loss thus trades off having a large margin for each SVM with being close to the mean model.

\cite{Jacob2009} make the assumptions underlying cluster regularization more explicit by formalizing a cluster constraint on $A$ under the assumption that the number of clusters $C$ is known in advance. They then decompose the penalty into three separate norms:

\begin{itemize}
\item A global penalty which measures how large our column parameter vectors are on average: $\Omega_{mean}(A) = \|\bar{a}\|^2 $.
\item A measure of between-cluster variance that measures how close to each other the clusters are:  $\Omega_{between}(A) = \sum^C_{c=1} T_c \| \bar{a}_c - \bar{a} \|^2 $ where $T_c$ is the number of tasks in the $c$-th cluster and $\bar{a}_c$ is the mean vector of the task parameter vectors in the $c$-th cluster.
\item A measure of within-cluster variance that gauges how compact each cluster is: $\Omega_{within} = \sum^C_{c=1} \sum_{t \in J(c)} \| a_{\cdot, t} - \bar{a}_c \| $ where $J(c)$ is the set of tasks in the $c$-th cluster.
\end{itemize}

The final constraint then is the weighted sum of the three norms:
$$\Omega(A) = \lambda_1 \Omega_{mean}(A) + \lambda_2 \Omega_{between}(A) + \lambda_3 \Omega_{within}(A)$$
As this constraint assumes clusters are known in advance, they introduce a convex relaxation of the above penalty that allows to learn the clusters at the same time.

In another scenario, tasks might not occur in clusters but have an inherent structure. \cite{Kim2010} extend the group lasso to deal with tasks that occur in a tree structure, while \cite{Chen2010} apply it to tasks with graph structures.

While the previous approaches to modelling the relationship between tasks employ norm regularization, other approaches do so without regularization: \cite{Thrun1996} were the first ones who presented a task clustering algorithm using k-nearest neighbour, while \cite{Ando2005} learn a common structure from multiple related tasks with an application to semi-supervised learning.

Much other work on learning task relationships for multi-task learning uses Bayesian methods: 
\cite{Heskes2000} propose a Bayesian neural network for multi-task learning by placing a prior on the model parameters to encourage similar parameters across tasks. \cite{Lawrence2004} extend Gaussian processes (GP) to MTL by inferring parameters for a shared covariance matrix. As this is computationally very expensive, they adopt a sparse approximation scheme that greedily selects the most informative examples. \cite{Yu2005} also use GP for MTL by assuming that all models are sampled from a common prior. 

\cite{Bakker2003} place a Gaussian as a prior distribution on each task-specific layer. In order to encourage similarity between different tasks, they propose to make the mean task-dependent and introduce a clustering of the tasks using a mixture distribution. Importantly, they require task characteristics that define the clusters and the number of mixtures to be specified in advance.

Building on this, \cite{Xue2007} draw the distribution from a Dirichlet process and enable the model to learn the similarity between tasks as well as the number of clusters. They then share the same model among all tasks in the same cluster. \cite{DaumeIII2009} propose a hierarchical Bayesian model, which learns a latent task hierarchy, while \cite{Zhang2010a} use a GP-based regularization for MTL and extend a previous GP-based approach to be more computationally feasible in larger settings. 

Other approaches focus on the online multi-task learning setting: \cite{Cavallanti2010} adapt some existing methods such as the approach by \cite{Evgeniou2005} to the online setting. They also propose a MTL extension of the regularized Perceptron, which encodes task relatedness in a matrix. They use different forms of regularization to bias this task relatedness matrix, e.g. the closeness of the task characteristic vectors or the dimension of the spanned subspace. Importantly, similar to some earlier approaches, they require the task characteristics that make up this matrix to be provided in advance. \cite{Saha2011} then extend the previous approach by learning the task relationship matrix.

\cite{Kang2011} assume that tasks form disjoint groups and that the tasks within each group lie in a low-dimensional subspace. Within each group, tasks share the same feature representation whose parameters are learned jointly together with the group assignment matrix using an alternating minimization scheme. However, a total disjointness between groups might not be the ideal way, as the tasks might still share some features that are helpful for prediction. 

\cite{Kumar2012} in turn allow two tasks from different groups to overlap by assuming that there exist a small number of latent basis tasks. They then model the parameter vector $a_t$ of every actual task $t$ as a linear combination of these: $a_t = Ls_t$ where $L \in \mathbb{R}^{k \times d}$ is a matrix containing the parameter vectors of $k$ latent tasks, while $s_t \in \mathbb{R}^k$ is a vector containing the coefficients of the linear combination. In addition, they constrain the linear combination to be sparse in the latent tasks; the overlap in the sparsity patterns between two tasks then controls the amount of sharing between these. Finally, \cite{Crammer2012} learn a small pool of shared hypotheses and then map each task to a single hypothesis.

\section{Recent work on MTL for Deep Learning} \label{sec:recent_work}

While many recent Deep Learning approaches have used multi-task learning -- either explicitly or implicitly -- as part of their model (prominent examples will be featured in the next section), they all employ the two approaches we introduced earlier, hard and soft parameter sharing. In contrast, only a few papers have looked at developing better mechanisms for MTL in deep neural networks.

\subsection{Deep Relationship Networks}

\begin{figure}[!htb]
      \centering
         \includegraphics[width=1.0 \linewidth]{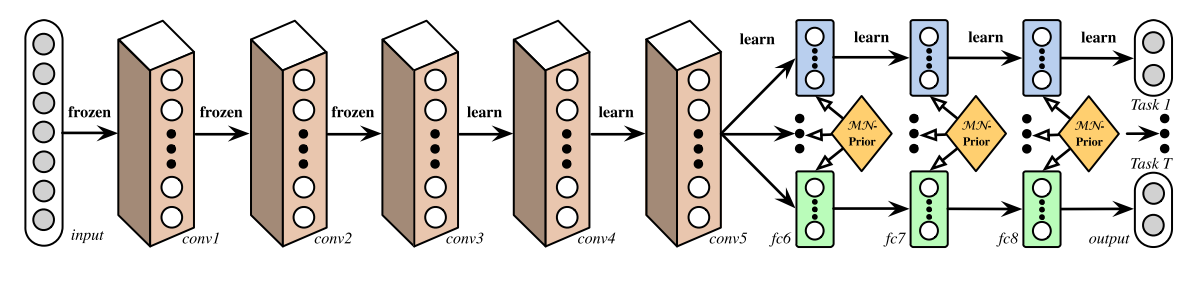}
	\caption{A Deep Relationship Network with shared convolutional and task-specific fully connected layers with matrix priors \cite{Long2015}}
	\label{fig:relationship_network}
\end{figure}

In MTL for computer vision, approaches often share the convolutional layers, while learning task-specific fully-connected layers. \cite{Long2015} improve upon these models by proposing Deep Relationship Networks. In addition to the structure of shared and task-specific layers, which can be seen in Figure \ref{fig:relationship_network}, they place matrix priors on the fully connected layers, which allow the model to learn the relationship between tasks, similar to some of the Bayesian models we have looked at before. This approach, however, still relies on a pre-defined structure for sharing, which may be adequate for well-studied computer vision problems, but prove error-prone for novel tasks.

\subsection{Fully-Adaptive Feature Sharing}

\begin{figure}[!htb]
      \centering
         \includegraphics[width=1.0 \linewidth]{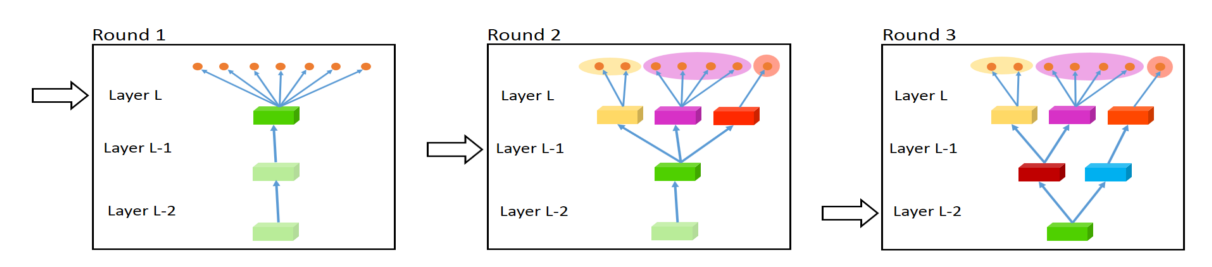}
	\caption{The widening procedure for fully-adaptive feature sharing \cite{Lu2016c}}
	\label{fig:fully-adaptive_feature_sharing}
\end{figure}

Starting at the other extreme, \cite{Lu2016c} propose a bottom-up approach that starts with a thin network and dynamically widens it greedily during training using a criterion that promotes grouping of similar tasks. The widening procedure, which dynamically creates branches can be seen in Figure \ref{fig:fully-adaptive_feature_sharing}. However, the greedy method might not be able to discover a model that is globally optimal, while assigning each branch to exactly one task does not allow the model to learn more complex interactions between tasks.

\subsection{Cross-stitch Networks}

\begin{figure}[!htb]
      \centering
         \includegraphics[width=0.60 \linewidth]{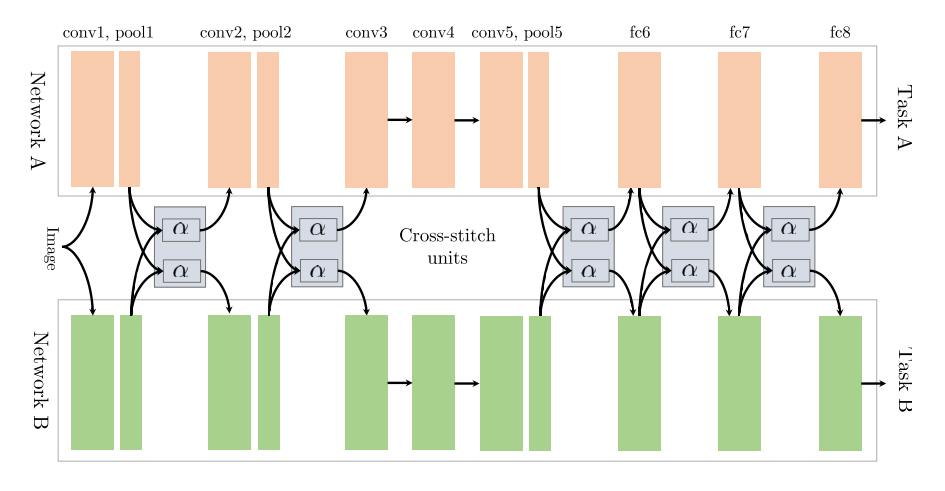}
	\caption{Cross-stitch networks for two tasks \cite{Misra2016}}
	\label{fig:cross-stitch_networks}
\end{figure}

\cite{Misra2016} start out with two separate model architectures just as in soft parameter sharing. They then use what they refer to as cross-stitch units to allow the model to determine in what way the task-specific networks leverage the knowledge of the other task by learning a linear combination of the output of the previous layers. Their architecture can be seen in Figure \ref{fig:cross-stitch_networks}, in which they only place cross-stitch units after pooling and fully-connected layers.

\subsection{Low supervision}

In contrast, in natural language processing (NLP), recent work focused on finding better task hierarchies for multi-task learning: \cite{Sogaard2016} show that low-level tasks, i.e. NLP tasks typically used for preprocessing such as part-of-speech tagging and named entity recognition, should be supervised at lower layers when used as auxiliary task.

\subsection{A Joint Many-Task Model}

\begin{figure}[!htb]
      \centering
         \includegraphics[width=0.5 \linewidth]{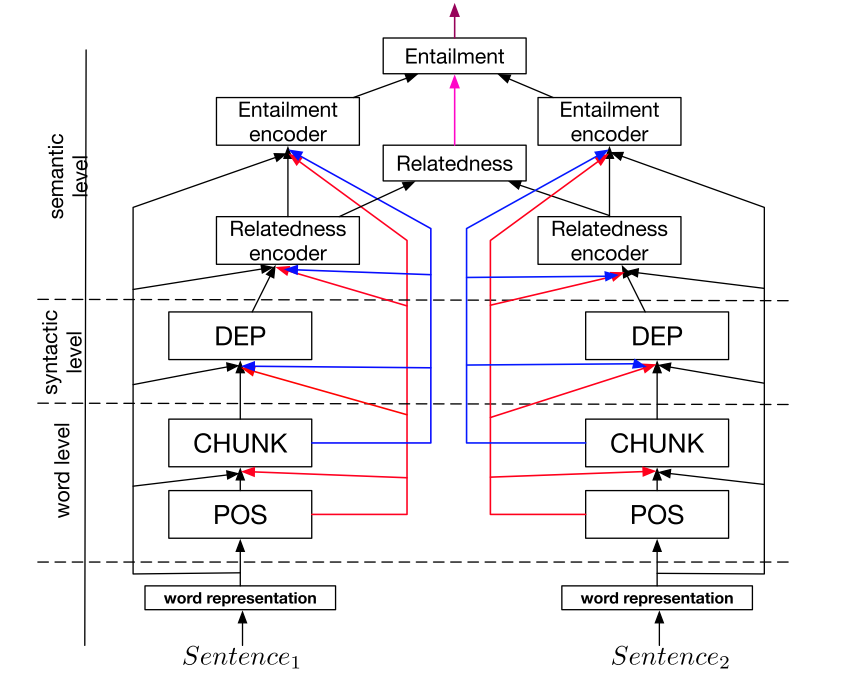}
	\caption{A Joint Many-Task Model \cite{Hashimoto2016c}}
	\label{fig:joint_many-task}
\end{figure}

Building on this finding, \cite{Hashimoto2016c} pre-define a hierarchical architecture consisting of several NLP tasks, which can be seen in Figure \ref{fig:joint_many-task}, as a joint model for multi-task learning.

\subsection{Weighting losses with uncertainty}

\begin{figure}[!htb]
      \centering
         \includegraphics[width=0.8 \linewidth]{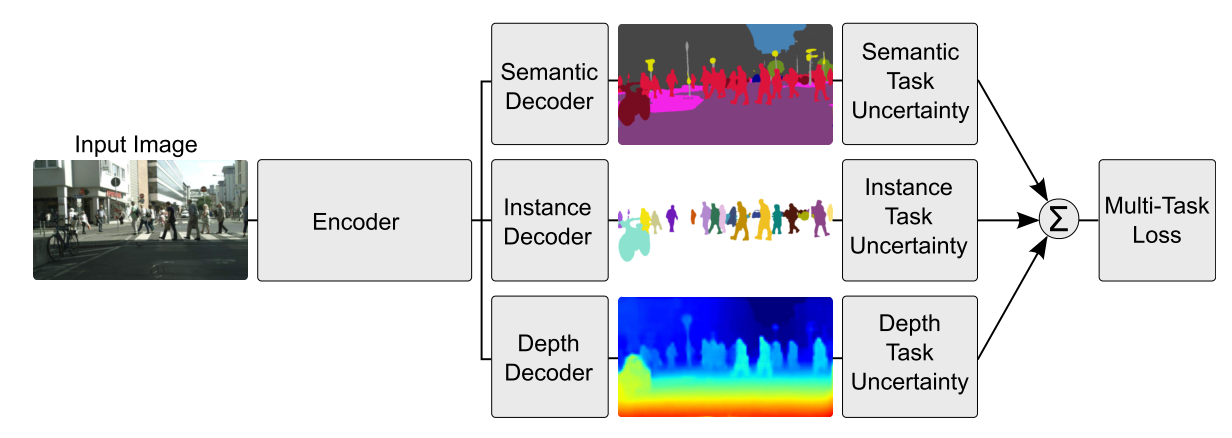}
	\caption{Uncertainty-based loss function weighting for multi-task learning \cite{Kendall2017}}
	\label{fig:uncertainty_weighting}
\end{figure}

Instead of learning the structure of sharing, \cite{Kendall2017} take an orthogonal approach by considering the uncertainty of each task. They then adjust each task's relative weight in the cost function by deriving a multi-task loss function based on maximizing the Gaussian likelihood with task-dependant uncertainty. Their architecture for per-pixel depth regression, semantic and instance segmentation can be seen in Figure \ref{fig:uncertainty_weighting}.

\subsection{Tensor factorisation for MTL}

More recent work seeks to generalize existing approaches to MTL to Deep Learning: \cite{Yang2017c} generalize some of the previously discussed matrix factorisation approaches using tensor factorisation to split the model parameters into shared and task-specific parameters for every layer.

\subsection{Sluice Networks}

\begin{figure}[!htb]
      \centering
         \includegraphics[width=0.60 \linewidth]{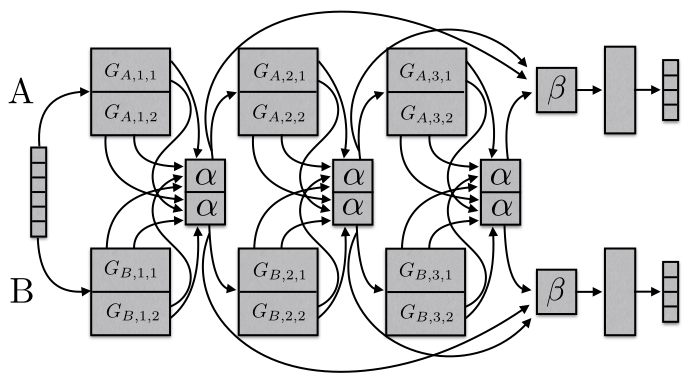}
	\caption{A sluice network for two tasks \cite{Ruder2017c}}
	\label{fig:sluice_network}
\end{figure}

Finally, we propose Sluice Networks \cite{Ruder2017c}, a model that generalizes Deep Learning-based MTL approaches such as hard parameter sharing and cross-stitch networks, block-sparse regularization approaches, as well as recent NLP approaches that create a task hierarchy. The model, which can be seen in Figure \ref{fig:sluice_network}, allows to learn what layers and subspaces should be shared, as well as at what layers the network has learned the best representations of the input sequences.

\subsection{What should I share in my model?}

Having surveyed these recent approaches, let us now briefly summarize and draw a conclusion on what to share in our deep MTL models. Most approaches in the history of MTL have focused on the scenario where tasks are drawn from the same distribution \cite{Baxter1997}. While this scenario is beneficial for sharing, it does not always hold. In order to develop robust models for MTL, we thus have to be able to deal with unrelated or only loosely related tasks.

While early work in MTL for Deep Learning has pre-specified which layers to share for each task pairing, this strategy does not scale and heavily biases MTL architectures. Hard parameter sharing, a technique that was originally proposed by \cite{Caruana1993}, is still the norm 20 years later. While useful in many scenarios, hard parameter sharing quickly breaks down if tasks are not closely related or require reasoning on different levels. Recent approaches have thus looked towards \emph{learning} what to share and generally outperform hard parameter sharing. In addition, giving our models the capacity to learn a task hierarchy is helpful, particularly in cases that require different granularities.

As mentioned initially, we are doing MTL as soon as we are optimizing more than one loss function. Rather than constraining our model to compress the knowledge of all tasks into the same parameter space, it is thus helpful to draw on the advances in MTL that we have discussed and enable our model to learn how the tasks should interact with each other.

\section{Auxiliary tasks} \label{sec:auxiliary_tasks}

MTL is a natural fit in situations where we are interested in obtaining predictions for multiple tasks at once. Such scenarios are common for instance in finance or economics forecasting, where we might want to predict the value of many possibly related indicators, or in bioinformatics where we might want to predict symptoms for multiple diseases simultaneously. In scenarios such as drug discovery, where tens or hundreds of active compounds should be predicted, MTL accuracy increases continuously with the number of tasks \cite{Ramsundar2015}.

In most situations, however, we only care about performance on one task. In this section, we will thus look at how we can find a suitable auxiliary task in order to still reap the benefits of multi-task learning. 

\subsection{Related task}

Using a related task as an auxiliary task for MTL is the classical choice. To get an idea what a related task can be, we will present some prominent examples. \cite{Caruana1998} uses tasks that predict different characteristics of the road as auxiliary tasks for predicting the steering direction in a self-driving car; \cite{Zhang2014b} use head pose estimation and facial attribute inference as auxiliary tasks for facial landmark detection; \cite{Liu2015a} jointly learn query classification and web search; \cite{Girshick2015} jointly predicts the class and the coordinates of an object in an image; finally, \cite{Ark2017} jointly predict the phoneme duration and frequency profile for text-to-speech.

\subsection{Adversarial}

Often, labeled data for a related task is unavailable. In some circumstances, however, we have access to a task that is \emph{opposite} of what we want to achieve. This data can be leveraged using an adversarial loss, which does not seek to minimize but maximize the training error using a gradient reversal layer. This setup has found recent success in domain adaptation \cite{Lempitsky2015}. The adversarial task in this case is predicting the domain of the input; by reversing the gradient of the adversarial task, the adversarial task loss is maximized, which is beneficial for the main task as it forces the model to learn representations that cannot distinguish between domains.

\subsection{Hints}

As mentioned before, MTL can be used to learn features that might not be easy to learn just using the original task. An effective way to achieve this is to use hints, i.e. predicting the features as an auxiliary task. Recent examples of this strategy in the context of natural language processing are \cite{Yu2016a} who predict whether an input sentence contains a positive or negative sentiment word as auxiliary tasks for sentiment analysis and \cite{Cheng2015} who predict whether a name is present in a sentence as auxiliary task for name error detection.

\subsection{Focusing attention}

Similarly, the auxiliary task can be used to focus attention on parts of the image that a network might normally ignore. For instance, for learning to steer \cite{Caruana1998} a single-task model might typically ignore lane markings as these make up only a small part of the image and are not always present. Predicting lane markings as auxiliary task, however, forces the model to learn to represent them; this knowledge can then also be used for the main task. Analogously, for facial recognition, one might learn to predict the location of facial landmarks as auxiliary tasks, since these are often distinctive.

\subsection{Quantization smoothing}

For many tasks, the training objective is quantized, i.e. while a continuous scale might be more plausible, labels are available as a discrete set. This is the case in many scenarios that require human assessment for data gathering, such as predicting the risk of a disease (e.g. low/medium/high) or sentiment analysis (positive/neutral/negative). Using less quantized auxiliary tasks might help in these cases, as they might be learned more easily due to their objective being smoother. 

\subsection{Predicting inputs}

In some scenarios, it is impractical to use some features as inputs as they are unhelpful for predicting the desired objective. However, they might still be able to guide the learning of the task. In those cases, the features can be used as outputs rather than inputs. \cite{Caruana1997} present several problems where this is applicable.

\subsection{Using the future to predict the present}

In many situations, some features only become available \emph{after} the predictions are supposed to be made. For instance, for self-driving cars, more accurate measurements of obstacles and lane markings can be made once the car is passing them. \cite{Caruana1998} also gives the example of pneumonia prediction, after which the results of additional medical trials will be available. For these examples, the additional data cannot be used as features as it will not be available as input at runtime. However, it can be used as an auxiliary task to impart additional knowledge to the model during training.

\subsection{Representation learning}

The goal of an auxiliary task in MTL is to enable the model to learn representations that are shared or helpful for the main task. All auxiliary tasks discussed so far do this implicitly: They are closely related to the main task, so that learning them likely allows the model to learn beneficial representations. A more explicit modelling is possible, for instance by employing a task that is known to enable a model to learn transferable representations. The language modelling objective as employed by \cite{Cheng2015} and \cite{Rei2017} fulfils this role. In a similar vein, an autoencoder objective can also be used as an auxiliary task.

\subsection{What auxiliary tasks are helpful?}

In this section, we have discussed different auxiliary tasks that can be used to leverage MTL even if we only care about one task. We still do not know, though, what auxiliary task will be useful in practice. Finding an auxiliary task is largely based on the assumption that the auxiliary task should be related to the main task in some way and that it should be helpful for predicting the main task.

However, we still do not have a good notion of when two tasks should be considered similar or related. \cite{Caruana1998} defines two tasks to be similar if they use the same features to make a decision. \cite{Baxter2000} argues only theoretically that related tasks share a common optimal hypothesis class, i.e. have the same inductive bias. \cite{Ben-David2003a} propose that two tasks are $\mathcal{F}$-related if the data for both tasks can be generated from a fixed probability distribution using a set of transformations $\mathcal{F}$. While this allows to reason over tasks where different sensors collect data for the same classification problem, e.g. object recognition with data from cameras with different angles and lighting conditions, it is not applicable to tasks that do not deal with the same problem. \cite{Xue2007} finally argue that two tasks are similar if their classification boundaries, i.e. parameter vectors are close.

In spite of these early theoretical advances in understanding task relatedness, we have not made much recent progress towards this goal. Task similarity is not binary, but resides on a spectrum. Allowing our models to learn what to share with each task might allow us to temporarily circumvent the lack of theory and make better use even of only loosely related tasks. However, we also need to develop a more principled notion of task similarity with regard to MTL in order to know which tasks we should prefer.

Recent work \cite{Alonso2017} has found auxiliary tasks with compact and uniform label distributions to be preferable for sequence tagging problems in NLP, which we have confirmed in experiments \cite{Ruder2017c}. In addition, gains have been found to be more likely for main tasks that quickly plateau with non-plateauing auxiliary tasks \cite{Bingel2017}. These experiments, however, have so far been limited in scope and recent findings only provide the first clues towards a deeper understanding of multi-task learning in neural networks. 

\section{Conclusion} \label{sec:conclusion}

In this overview, I have reviewed both the history of literature in multi-task learning as well as more recent work on MTL for Deep Learning. While MTL is being more frequently used, the 20-year old hard parameter sharing paradigm is still pervasive for neural-network based MTL. Recent advances on learning what to share, however, are promising. At the same time, our understanding of tasks -- their similarity, relationship, hierarchy, and benefit for MTL -- is still limited and we need to study them more thoroughly to gain a better understanding of the generalization capabilities of MTL with regard to deep neural networks.

\bibliography{multi-task}
\bibliographystyle{apalike}

\end{document}